\documentclass[runningheads]{llncs}

\usepackage[T1]{fontenc}
\usepackage{graphicx}
\graphicspath{{assets/}}
\usepackage{amsmath}
\usepackage{amssymb}
\usepackage{multirow}
\usepackage[table]{xcolor}%
\usepackage{eucal}%

\newcommand\pa[1]{\ensuremath{\left( #1 \right)}} %
\newcommand\pc[1]{\ensuremath{\left\{ #1 \right\}}} %
\newcommand\R{\ensuremath{\mathbb{R}}} %

\newcommand\sL{\ensuremath{\mathcal{L}}}
\newcommand\sU{\ensuremath{\mathcal{U}}}
\newcommand\sX{\ensuremath{\mathcal{X}}}
\newcommand\sY{\ensuremath{\mathcal{Y}}}
\newcommand\sP{\ensuremath{\mathcal{P}}} %
\newcommand{\norm}[1]{\left\lVert#1\right\rVert}

\DeclareMathOperator*{\argmax}{\arg\,\max}

\usepackage{array}
\newcolumntype{C}[1]{>{\arraybackslash\centering}p{#1}}
\newcolumntype{R}[1]{>{\arraybackslash\raggedleft}p{#1}}
\newcolumntype{L}[1]{>{\arraybackslash\raggedright}p{#1}}

\newsavebox\CBox 
\def\textBF#1{\sbox\CBox{#1}\resizebox{\wd\CBox}{\ht\CBox}{\textbf{#1}}}

\usepackage{tikz}
\newcommand{\ArbitraryLengthArrow}[1]{%
    \parbox{#1}{\tikz{\draw[->](0,0)--(#1,0);}}
}

\newcommand{\Ttablebottomvspace}{\vspace{0mm}}
\newcommand{\Tfigurebottomvspace}{\vspace{0mm}}

\begin{document}

\title{ Using Multiple Instance Learning to Build \\ Multimodal Representations  } %
 
\author{Peiqi Wang$^1$, William M. Wells$^{1}$, Seth Berkowitz$^2$, \\
        Steven Horng$^2$, and Polina Golland$^1$} %
\authorrunning{P. Wang et al.}
\institute{
    $^1$ CSAIL, MIT, Cambridge MA, USA  \\
    $^2$ BIDMC, Harvard Medical School, Boston, MA, USA \\
    \email{wpq@mit.edu, polina@csail.mit.edu}}

\maketitle              %
\begin{abstract}

Image-text multimodal representation learning aligns data across modalities and enables important medical applications, e.g., image classification, visual grounding, and cross-modal retrieval. In this work, we establish a connection between multimodal representation learning and multiple instance learning. Based on this connection, we propose a generic framework for constructing permutation-invariant score functions with many existing multimodal representation learning approaches as special cases. Furthermore, we use the framework to derive a novel contrastive learning approach and demonstrate that our method achieves state-of-the-art results in several downstream tasks. 

\end{abstract}

\section{Introduction}

In this paper, we propose a framework for designing multimodal representation learning methods that encompasses previous approaches as special cases and implies a new algorithm for multimodal learning that advances the state of the art. Specifically, we establish a connection between self-supervised representation learning based on contrastive learning and multiple instance learning \cite{carbonneauMultipleInstanceLearning2018} and show that they share similar assumptions and goals. We bring insights from multiple instance learning to offer a fresh perspective on self-supervised representation learning and ideas for performance improvements. With this connection in mind, we derive a novel algorithm for learning image-text representations that capture the structure shared between the two modalities and generalize well in a variety of downstream tasks.

We aim to establish alignment between images and associated text to improve clinical workflow. For example, an image model that mimics the radiologists' interpretation could retroactively label images to select relevant patients for a clinical trial. Further, local alignment between image regions and text fragments (e.g., sentences) promises to benefit many downstream tasks. For example, cross-modal retrieval can provide description of an image region for automated documentation or enable comparisons with similar previously imaged patients for better interpretation based on local anatomy or pathology. Similarly, radiologists documenting findings can verify the accuracy of the report by noting if the referred location (i.e., visual grounding of the text) is consistent with their impression of the image.

Self-supervised representation learning is a useful tool for reducing annotation burden for machine learning models in medical imaging. Despite the need and opportunities for automation, development of robust machine learning methods is held back by the lack of annotations that serve as the supervision signal for learning. Self-supervised representation learning on paired image-text data offers two advantages: (i) learning requires no further annotations and (ii) treating text as ``labels'' enables us to use natural language to reference visual concepts and vice versa \cite{radfordLearningTransferableVisual2021}. Thus, we focus on learning image-text multimodal representations but the proposed framework is broadly applicable to representation learning on other multimodal data.

Learning joint representations involves training image and text encoders to perform self-supervised tasks on paired image-text data \cite{chenUNITERUNiversalImageTExt2020a,liVisualBERTSimplePerformant2019,luViLBERTPretrainingTaskAgnostic2019} and evaluating on relevant downstream tasks. We focus on contrastive learning, i.e., classifying image-text pairs as matched (i.e., corresponding to the same imaging event), or mismatched. Contrastive learning has been applied to the medical domain, demonstrating impressive transfer capabilities on a diverse set of tasks \cite{boeckingMakingMostText2022,chauhanJointModelingChest2020,huangGLoRIAMultimodalGlobalLocal2021,liaoMultimodalRepresentationLearning2021,mullerJointLearningLocalized2022,zhangContrastiveLearningMedical2022}. The biggest improvements come from addressing challenges unique to this domain, e.g., the use of cross attention to deal with the lack of effective pathology detectors \cite{huangGLoRIAMultimodalGlobalLocal2021} and adaptation of language models to address linguistic challenges in clinical notes \cite{boeckingMakingMostText2022}. Training the models has involved increasingly complex contrastive loss functions that treat image and text symmetrically \cite{boeckingMakingMostText2022,chauhanJointModelingChest2020,huangGLoRIAMultimodalGlobalLocal2021,zhangContrastiveLearningMedical2022} and on multiple scales \cite{boeckingMakingMostText2022,huangGLoRIAMultimodalGlobalLocal2021,liaoMultimodalRepresentationLearning2021,mullerJointLearningLocalized2022}. In contrast to previous work that relies on many loss terms, our proposed contrastive loss is simple to implement and yields superior performance.

Borrowing ideas from multiple instance learning, we treat local image region features as ``data'' and sentence features as (complex) ``labels''. Multiple instance learning is a type of weakly supervised learning that is effective for problems that lack fine-grain annotations \cite{carbonneauMultipleInstanceLearning2018}. For example, it can help to locate tumor cells in whole slide images with just image-level labels \cite{liDualstreamMultipleInstance2021}. Central to multiple instance learning is the construction of permutation-invariant score functions \cite{ilseAttentionbasedDeepMultiple2018}, and the choice of how the instance scores or features are aggregated to be evaluated against an image-level label. Effective instance aggregators leverage domain knowledge \cite{fouldsReviewMultiinstanceLearning2010}, e.g., the Noisy-OR aggregator for drug activity prediction \cite{maronFrameworkMultipleInstanceLearning1998}, the Noisy-AND aggregator for cellular phenotype classification \cite{krausClassifyingSegmentingMicroscopy2016a}. In our work, we extend multiple instance classification to contrastive learning by constructing permutation-invariant image-text score functions. Drawing on insights from multiple instance classification with correlated instances \cite{liDualstreamMultipleInstance2021}, our proposed instance aggregator exploits correlation among instances to build representations that perform well in downstream tasks.

Many prior multiple instance learning methods focused on one particular task of interest, e.g., detection \cite{zhangMultipleInstanceBoosting2005}, region classification \cite{fangCaptionsVisualConcepts2015}, or retrieval \cite{karpathyDeepFragmentEmbeddings2014}. Some investigated the choices of instance aggregators for more than one downstream task \cite{harwathJointlyDiscoveringVisual2020,miechEndtoEndLearningVisual2020a} but are limited in generality (i.e., not intended for other applications) and scope (i.e., explored a few simple instance aggregators). In contrast, our proposed framework for constructing permutation-invariant score functions can be readily applied to other applications. We systematically investigate instance aggregators and their effect on representation learning, leading to a novel approach for learning joint representations. We evaluate the resulting image-text representations on a diverse set of downstream tasks and demonstrate state-of-the-art performance across all tasks in the context of a large set of chest X-ray images and associated radiological reports.

\section{Method}
\label{sec:method}

We first introduce notation and discuss the local and global approaches for constructing permutation-invariant image-document score functions at the core of the learning procedure. We then instantiate the framework for a specific choice of aggregators for contrastive learning.

\subsection{Problem Setup}

A local $D$-dimensional representation of an image with $N$ proposed regions is a collection of $N$ features vectors $x_n\in \sX \subset \R^D$, $n\in\pc{1,\cdots,N}$. In our experiments, we use regular tiling to generate image regions and leave more sophisticated proposal methods (e.g., \cite{renFasterRCNNRealTime2015}) for future work. A local representation of a $M$-sentence document (e.g., a radiology report) is a collection of sentence feature vectors $y_m \in\sY\subset \R^D$, $m\in\pc{1,\cdots,M}$.

Function $h: \sX\times \sY \to\R$ measures the similarity between representations, e.g., $h(x_n,y_m)$ is the similarity between a region and a sentence. In our experiments, we use cosine similarity $h(x, y) = \langle x, y \rangle / \pa{\norm{x} \norm{y}}$, though the formulation accepts any differentiable similarity function. 

For any vector space $\sU$, aggregator function $\pi: \sP(\sU) \to \sU$ aggregates elements in the input set into a ``representative''. $\sP(\sU)$ is the set of all finite subsets of $\sU$. For example, $\pi(\pc{x_n}) = \frac{1}{N}\sum_n x_n$ aggregates $N$ region features $x_n\in\sX$ by averaging them, while $\pi(\pc{h_n}) = \max_n h_n$ aggregates $N$ similarity scores into a single score by computing the maximum score. We restrict our attention to aggregators that are permutation-invariant, i.e., they treat their input as an unordered set rather than an ordered vector. 

Permutation-invariant image-document score function $S: \sP(\sX) \times \sP(\sY) \to\R$ measures the similarity between an image and a document based on region features $\pc{x_n}$ and sentence features $\pc{y_m}$. 

\begin{figure}[ht]
\centering
\includegraphics[width=\linewidth]{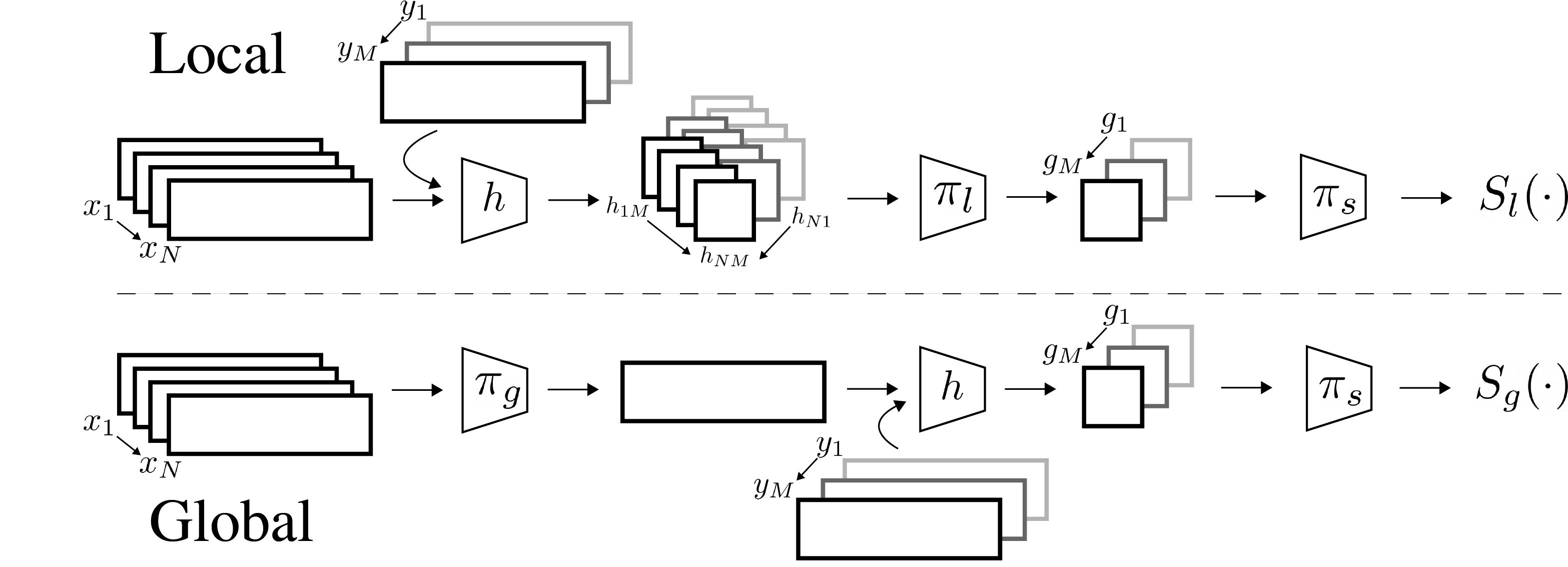}
\caption{\footnotesize Local (top) and global (bottom) image-document score functions.}
\label{fig:score_function_schematics}
\Tfigurebottomvspace
\end{figure}

\subsection{Local \& Global Permutation-Invariant Score Functions}
\label{sec:perm_inv_score_functions}

Contrastive representation learning can be seen as maximizing the likelihood of correctly classifying image-text pairs as matched or mismatched. Since supervision is provided at the image-document level, we define a framework to build permutation-invariant image-document score functions.

{\bfseries The local approach} aggregates region-sentence scores into an image-sentence score. The image-sentence score $g_m$ for sentence $m$ in the document is obtained by applying a local aggregator function $\pi_l$ to region-sentence scores, i.e., $g_m = \pi_l ( \pc{h(x_n, y_m)}_n ) \triangleq \pi_l ( \pc{h(x_1, y_m), \cdots, h(x_N, y_m)} ) $. 

{\bfseries The global approach} first aggregates local region features $\pc{x_n}$ into a single image feature vector $\pi_g(\pc{x_n})$ using a global aggregator function $\pi_g$. The image-sentence score $g_m$ is computed using the similarity function $h$ on the image feature vector $\pi_g(\pc{x_n})$ and sentence feature vector $y_m$, i.e., $g_m = h(\pi_g(\pc{x_n}), y_m)$. 

In both approaches, the image-document score $S$ is obtained by aggregating image-sentence scores with another aggregator function $\pi_s$, i.e., $S(\pc{x_n}, \pc{y_m}) = \pi_s(\pc{g_m})$. Figure~\ref{fig:score_function_schematics} illustrates the framework for constructing $S$. To summarize, the local and global image-document scores $S_l$ and $S_g$ are computed as follows:
\begin{align}
    \label{eq:method_S_l}
    S_l(\pc{x_n}, \pc{y_m})
        &= \pi_s(\pc{ \pi_l(\pc{ h(x_n, y_m) }_n) }_m), \\ 
    \label{eq:method_S_g}
    S_g(\pc{x_n}, \pc{y_m})
        &= \pi_s(\pc{ h( \pi_g(\pc{x_n}), y_m ) }_m).
\end{align}

As the aggregator functions are permutation-invariant, the image-document score function $S$ is naturally permutation-invariant as well. We emphasize that $S$ treats image features and text features differently, and that the order of application of similarity evaluation $h(\cdot)$ and aggregators $\pi(\cdot)$ is empirically relevant. This design decision is motivated by the fact that each sentence in a radiology report represent a concept and its location in the image, i.e., it is akin to a label for some region in the image. The converse is not necessarily true as some parts of the image are not described in the report.

\subsection{Representation Learning with LSE$+$NL Aggregators}

In this section, we introduce our method LSE$+$NL for learning multimodal representations that relies on a combination of local and global image-document score functions and an asymmetric text-to-image contrastive loss.

Inspired by \cite{liDualstreamMultipleInstance2021}, we use a soft maximum function to identify the most relevant region for a sentence, i.e., the critical region, and attend more to regions that are similar to the critical region. Specifically, the local aggregator $\pi_l$ is the log-sum-exp (LSE) function
\begin{align}
    \pi_l(\pc{h_n}) 
        = \frac{1}{\gamma_l} \log\sum_{n=1}^N \exp(\gamma_l\,h_n),
    \label{eq:agg_lse}
\end{align}
where $\gamma_l$ is a scale parameter that controls how well the LSE function approximates the max function. The global aggregator $\pi_g$ linearly combines the region features using the distance to the critical region as weights, i.e., 
\begin{align}
    \pi_g(\pc{x_n}) 
        = \sum_{n=1}^N  \dfrac{ \exp( \gamma_g\, \langle Ax_n, Ax_k \rangle) }{ \sum_{n'=1}^N \exp(\gamma_g\, \langle Ax_{n'}, Ax_k \rangle) } \, x_n,
    \label{eq:agg_nonlocal_attention}
\end{align} 
where $k$ is the index of the critical region, i.e., $k=\argmax_n h(x_n,y_m)$, $A$ is a learned weight matrix, and $\gamma_g$ is the scale parameter for the softmax function. We can interpret $\pi_g$ as a form of attention where regions that are more similar to the critical region are given a higher attention weight. In effect, $\pi_g$ exploits the correlation between each region and the critical region using attention. In addition, $\pi_g$ can be seen as a form of non-local (NL) network \cite{wangNonlocalNeuralNetworks2018}. Both $\pi_l$ and $\pi_g$ are permutation-invariant functions. We choose $\pi_s$ to be the average function.

We use the local and global image-document scores in (\ref{eq:method_S_l}) and (\ref{eq:method_S_g}) computed with our choice of $\pi_l$ and $\pi_g$ for contrastive learning. Given a document, we form an image-document score vector $s \triangleq (s^+,s^-_1,\cdots,s^-_K)$ where $s^+\in\R$ is the image-document score with its matched image and $s^-_k\in\R$ for $k=1,\cdots,K$ is the image-document score with $K$ mismatched images. We use $s_l$ and $s_g$ to denote $(K+1)$-length score vectors defined above computed using the local and the global score functions respectively. The image and text encoders are trained to minimize $\sL(s_l)+\sL(s_g)$ over documents in the training set where $\sL$ is the text-to-image contrastive loss \cite{oordRepresentationLearningContrastive2018,zhangContrastiveLearningMedical2022}
\begin{align}
    \sL(s)
        \triangleq -\log \frac{ \exp(\gamma\,s^+) }{ \exp(\gamma\,s^+) + \sum_{k=1}^K \exp(\gamma\,s^-_k) }
    \label{eq:infonce_loss}
\end{align} 
with scale parameter $\gamma$. In the equation above, $s$ is either vector $s_l$ computed using (\ref{eq:method_S_l}) with $\pi_l$ defined in (\ref{eq:agg_lse}) or vector $s_g$ computed using (\ref{eq:method_S_g}) with $\pi_g$ defined in (\ref{eq:agg_nonlocal_attention}). The image-to-text contrastive loss where the negative scores are computed for an image with $K$ different mismatched documents is often used alongside $\sL$ in prior work \cite{boeckingMakingMostText2022,chauhanJointModelingChest2020,huangGLoRIAMultimodalGlobalLocal2021,zhangContrastiveLearningMedical2022}. We choose to treat images and text asymmetrically and show that the simple text-to-image contrastive loss is sufficient to induce representations that generalize well.

\section{Connection to Multiple Instance Learning}

\newcommand{\Taverage}{Avg}
\newcommand{\Tidentity}{Id}
\newcommand{\Tsum}{Sum}
\newcommand{\Tmax}{Max}
\newcommand{\Tnonlocal}{NL}
\newcommand{\Tlse}{LSE}
\newcommand{\Tlset}{$\text{LSE}_t$}
\newcommand{\Tlseo}{$\text{LSE}_1$}
\newcommand{\Tnn}{NN}
\newcommand{\Tcrossattention}{CA}
\newcommand{\Tmissing}{-}

\newcommand{\Tvideo     }{video}
\newcommand{\Timage     }{image}
\newcommand{\Taudio     }{audio}
\newcommand{\Tregion    }{region}
\newcommand{\Tsentence  }{sentence}
\newcommand{\Tword      }{word}

\renewcommand{\arraystretch}{1.1}
\begin{table}[ht!]
\footnotesize
\centering
\caption{
Taxonomy of related methods for image-language representation learning in our multiple instance learning inspired framework. For each method, we report image segments captured by $x_n$ (\Tregion{} or \Tvideo{}), language segments captured by $y_m$ (\Tword{}, \Tsentence{}, or \Taudio{}), local aggregator $\pi_l$ if used (\Tmax{} or \Tlse{}), global aggregator $\pi_g$ if used (\Taverage{}, \Tnn{} for generic non-linear functions, cross attention (\Tcrossattention{}) $\pi_l(\{x_n\},y_m) = \sum_n \exp(\langle x_n,y_m\rangle)/\sum_{n'} \exp(\langle x_{n'},y_m \rangle) x_n$, or \Tnonlocal{} in (\ref{eq:agg_nonlocal_attention})), and the final score aggregator $\pi_s$ (\Tsum{}, \Tmax{}, \Tlse{}, \Tidentity{}, \Taverage{}). 
}
\label{tab:pretrain_methods_under_mil_framework}
\begin{tabular}{L{.28\linewidth}C{.12\linewidth}C{.13\linewidth}C{.11\linewidth}C{.13\linewidth}C{.15\linewidth}p{0pt}}
\hline
Methods & $x_n$ & $y_m$ & $\pi_l$ & $\pi_g$ & $\pi_s$ & \\
\hline
NeuralTalk \cite{karpathyDeepVisualSemanticAlignments2017}                                      & \Tregion{} & \Tword{}      & \Tmax{}                      &  \Tmissing{}                & \Tsum{}                & \\
DAVEnet-MISA \cite{harwathJointlyDiscoveringVisual2020}                                         & \Tregion{} & \Taudio{}     & \Tmax{}                      &  \Tmissing{}                & \Tsum{}                & \\
MIML \cite{gaoLearningSeparateObject2018}                                                       & \Tvideo{}  & \Taudio{}     & \Tmax{}                      &  \Tmissing{}                & \Tmax{}                & \\ 
MIL-NCE \cite{miechEndtoEndLearningVisual2020a}                                                 & \Tvideo{}  & \Tsentence{}  & \Tmissing{}                  &  \Taverage{}                & \Tlse{}                & \\
ConVIRT/CLIP \cite{zhangContrastiveLearningMedical2022,radfordLearningTransferableVisual2021}   & \Tregion{} & \Tsentence{}  & \Tmissing{}                  &  \Tnn{} $\circ$ \Taverage{} & \Tidentity{}           & \\
GLoRIA/BioViL \cite{huangGLoRIAMultimodalGlobalLocal2021,boeckingMakingMostText2022}            & \Tregion{} & \Tword{}      & \Tmissing{}                  &  \Tcrossattention{}         & \Tlse{}                & \\
                                                                                                & \Tregion{} & \Tsentence{}  & \Tmissing{}                  &  \Taverage{}                & \Tidentity{}           & \\
\hline
LSE$+$NL (Ours)                                                                                 & \Tregion{} & \Tsentence{}  & \Tlse {}                     &  \Tmissing{}               & \Taverage{}            & \\
                                                                                                & \Tregion{} & \Tsentence{}  & \Tmissing{}                  &  \Tnonlocal{}              & \Taverage{}            & \\
\hline
\end{tabular}
\Ttablebottomvspace
\end{table}

In multiple instance learning \cite{carbonneauMultipleInstanceLearning2018}, a set that contains many instances $\pc{x_1,\cdots,x_N}$ is referred to as a bag. The training set consists of bags and their associated bag labels $y$ while the instance labels are not provided. For binary bag labels, a positive bag is guaranteed to include at least one positive instance, while a negative bag includes no positive instances. The bag-level labels are used to train classifier to assign instance-level and bag-level labels in new, unseen bags.

Existing image-text representation learning algorithms that are either predictive \cite{desaiVirTexLearningVisual2021} or contrastive \cite{radfordLearningTransferableVisual2021} can be seen as a form of multiple instance learning. Specifically, we can view an image as a bag of region features and the corresponding sentence that describes the image as the bag label. Instead of taking on binary values, the bag labels can represent arbitrary categories via natural language. Although the exact region that corresponds to the sentence is unknown, the matched image contains at least one region that corresponds to the text while a randomly sampled image most likely does not. Similar to multiple instance learning, self-supervised representation learning methods use these assumptions for learning. 

More generally, we consider the text label as a bag of sentences. For example, sentences describing findings within a chest X-ray image most likely can be permuted without changing the overall meaning. Therefore, representation learning can be interpreted as predicting the label bag $\pc{y_m}$ given the input bag $\pc{x_m}$. This setup corresponds to multi-instance multi-label learning \cite{zhouMultiInstanceMultiLabelLearning2012}.

Moreover, multiple instance learning and multimodal representation learning share comparable goals. Multiple instance learning aims to align instances and bags with labels such that the pre-trained model performs well in classification tasks. Multimodal representation learning aims to align images and their subregions with text such that the pre-trained model perform well on tasks that rely on such alignment, e.g., image classification relies on image-sentence alignment, visual grounding and cross-modal retrieval rely on region-sentence alignment.

There are two main multiple instance learning approaches, instance-level and embedding-level approaches \cite{amoresMultipleInstanceClassification2013}. The instance-level approach computes the bag score by aggregating the instance scores, while the embedding-level approach computes the bag score based on a bag feature that is aggregated from the instance features. The local and global approaches in Section~\ref{sec:perm_inv_score_functions} are extensions of the instance and embedding approaches to contrastive learning.

This parallel enables us to analyze prior methods as instances of the framework defined in Section~\ref{sec:perm_inv_score_functions} that is inspired by multiple instance learning (Table~\ref{tab:pretrain_methods_under_mil_framework}). We make one generalization to the formulation in Section~\ref{sec:perm_inv_score_functions} to accommodate cross attention \cite{leeStackedCrossAttention2018}: the local aggregator function $\pi_l$ can potentially rely on label features $y_m$ to multiplex its behavior, i.e., $\pi_l: \sP(\sX)\times\sY \to\sX$. In summary, a diverse set of aggregators $\pi_l,\pi_g,\pi_s$ have been demonstrated on multimodal representation learning at varying scales, implying there may not be a single set of aggregators that works well for every problem. More realistically, the best aggregator functions are the ones that fit application-specific assumptions well.

\section{Experiments}

We illustrate the proposed approach by building a representation of frontal chest X-ray images and associated radiology reports and using it in downstream tasks. In all of the experiments, the data used for representation learning is disjoint from the test sets used to evaluate the downstream tasks.

We normalize the images and resize them to 512x512 resolution. We apply random image augmentations, i.e., 480x480 random crops, brightness and contrast variations, and random affine transforms (only for image model fine-tuning during evaluation). We use PySBD \cite{sadvilkarPySBDPragmaticSentence2020} for sentence tokenization. 

We employ ResNet-50 \cite{heDeepResidualLearning2016} as the image region encoder and CXR-BERT \cite{boeckingMakingMostText2022} as the sentence encoder. Each encoder is followed by a linear projection to a 128 dimension embedding space. In particular, the projected ResNet-50 conv-5 activations act as the region features $\pc{x_n}$ and the projected mean-pooled contextualized word embeddings acts as the sentence features $\pc{y_m}$.

\subsection{Representation learning}

We use a subset of 234,073 chest X-ray images and report from MIMIC-CXR \cite{johnsonMIMICCXRDeidentifiedPublicly2019a} for representation learning. We randomly initialize the image encoder and use the CXR-BERT model \cite{boeckingMakingMostText2022} pre-trained on a biomedical corpus (i.e., the stage II model) as the sentence encoder. We use the AdamW optimizer \cite{loshchilovDecoupledWeightDecay2019} and decay the initial learning rate of 5e-5 using a cosine schedule with 2k warmup steps. we initialize $\gamma$ to 14 and optimize this hyperparameter alongside the encoder parameters. We set other scale parameters as follows: $\gamma_l=0.1, \gamma_g=e$. We use a batch size of 64. For each image in the batch, we sample 5 sentences, with replacement if needed, to make up the label bag. Here, $N=225$ and $M=5$.

\subsection{Downstream Tasks}

\noindent{\bfseries Image Classification} To evaluate zero-shot (ZS) and fine-tuned (FT) classification performance, we use the same split of RSNA Pneumonia (RSNA) \cite{shihAugmentingNationalInstitutes2019} as in \cite{huangGLoRIAMultimodalGlobalLocal2021}, specifically, 18,678/4,003/4,003 for training/validation/testing. To evaluate in-distribution fine-tuned classification performance in the ablation study, we use 5 CheXpert labels (Atelectasis, Cardiomegaly, Edema, Pleural Effusion, Pneumothorax) on the MIMIC-CXR data set \cite{johnsonMIMICCXRDeidentifiedPublicly2019a} that we denote MIMIC-CheXpert (CheX). There are roughly 1k images in the test set associated with each CheXpert label. To evaluate the data efficiency of representation learning approaches, we use different amounts of training data (1\% and 100\%).

For zero-shot image classification, we first tokenize and encode the class-specific text prompts (e.g., ``Findings suggesting pneumonia.'' and ``No evidence of pneumonia.''). For each image, we assign a binary label that corresponds to the prompt with the higher image-sentence score. We find it important to normalize the scores to $[0,1]$ for each class before applying the softmax. For fine-tuned image classification, we use the Adam optimizer \cite{kingmaAdamMethodStochastic2014} with a learning rate of 3e-3 to optimize the randomly initialized weights and a bias over the mean-pooled region features while keeping the encoder weights fixed. For RSNA Pneumonia, we report accuracy and AUC. For MIMIC-CheXpert, we report the average AUC over five binary classification tasks.

\noindent{\bfseries Visual Grounding} We evaluate visual grounding performance using the MS-CXR region-sentence annotations \cite{boeckingMakingMostText2022}. This data set consists of 1,448 bounding boxes over 1,162 images, where each bounding box is associated with a sentence that describes its dominant radiological feature. We compute region-sentence scores to quantify how well the sentence is localized in the image. We report a measure of discrepancy between region-sentence scores inside and outside the bounding box, i.e., contrast-to-noise ratio (CNR) \cite{boeckingMakingMostText2022}, and how well the thresholded region-sentence scores overlap with the bounding box on average, i.e., mean intersection over union (mIoU). In contrast to \cite{boeckingMakingMostText2022}, we pick thresholds that span $[-1,1]$ in $0.05$ increments to compute the mIoU for a fair comparison.

\noindent{\bfseries Cross-Modal Retrieval} We evaluate cross-modal retrieval performance using the MS-CXR data set as well. We compute the bounding box features from the region features with RoIAlign \cite{heMaskRCNN2017}. We compute box-sentence scores and sort them to retrieve items in one modality given a query from the other modality. The correctly retrieved item is the one that is paired with the query item. We report the fraction of times the correct item was found in the top K results (R@K) and the median rank of the correct item in the ranked list (MedR).

\subsection{Results}

\renewcommand{\arraystretch}{1}
\begin{table}[ht!]
    \centering
    \caption{\footnotesize Image classification performance on the RSNA Pneumonia data set. We report accuracy and AUC on zero-shot and fine-tuned classification (fine-tuned on 1\% and 100\% labels). Our approach compares favorably to BioViL \cite{boeckingMakingMostText2022}. }
    \label{tab:exp_cmp_with_sota_rsna}
    \begin{tabular}{L{0.13\linewidth}  C{0.1\linewidth}C{0.1\linewidth} @{\hspace{10\tabcolsep}} C{0.1\linewidth}C{0.1\linewidth} @{\hspace{10\tabcolsep}} C{0.1\linewidth}C{0.1\linewidth}p{0pt}}
    \cline{1-8}
    Method & \multicolumn{2}{c}{Zero-Shot} & \multicolumn{2}{c}{1\%} & \multicolumn{2}{c}{100\%} &  \\
    & \scriptsize ACC$\uparrow$ & \scriptsize AUC$\uparrow$ & \scriptsize ACC$\uparrow$ & \scriptsize AUC$\uparrow$ & \scriptsize ACC$\uparrow$ & \scriptsize AUC$\uparrow$ & \\
    \cline{1-8}
    BioViL & 0.73 & 0.83 & 0.81 & \textBF{0.88} & 0.82 & \textBF{0.89} &\\
    LSE$+$NL & \textBF{0.80} & \textBF{0.84} & \textBF{0.84} & 0.87 & \textBF{0.85} & \textBF{0.89} &\\
    \cline{1-8}
    \end{tabular}
    \Ttablebottomvspace
\end{table}

\renewcommand{\arraystretch}{1}
\begin{table}[ht!]
    \begin{minipage}[c]{.65\textwidth}
        \caption{\footnotesize Visual grounding performance. We report contrast-to-noise ratio (CNR) and mean intersection-over-union (mIoU). mIoU measures mean IoU of a thresholded region-sentence map and the ground truth bounding box over a set of thresholds. Our approach outperforms BioViL \cite{boeckingMakingMostText2022} on both measures.
    }
    \label{tab:exp_cmp_with_sota_grounding}
    \end{minipage}
    \hfill
    \begin{minipage}[t]{.325\textwidth}
        \vspace{-13.5mm}
        \centering
        \begin{tabular}{L{0.35\linewidth}C{0.3\linewidth}C{0.3\linewidth}p{0pt}}
        \cline{1-3}
        Method & CNR$\uparrow$ & mIoU$\uparrow$ &  \\
        \cline{1-3}
        BioViL & 1.14 & 0.17 &  \\
        LSE$+$NL & \textBF{1.44} & \textBF{0.19} &  \\
        \cline{1-3}
        \end{tabular} 
        \end{minipage}
    \Ttablebottomvspace
    \vspace{-4mm}
\end{table} 

\renewcommand{\arraystretch}{1}
\begin{table}[ht!]
\centering
\caption{\footnotesize Cross-modal retrieval performance. We report recall for the top 10, 50 and 100 answers returned by the method, as well as the median rank of the ground truth element for sentence retrieval based on region queries and for region retrieval based on sentence queries. Our method outperforms the baselines on all measures. }
\label{tab:exp_cmp_with_sota_retrieval}
\begin{tabular}{L{0.13\linewidth} C{0.09\linewidth}C{0.09\linewidth}C{0.09\linewidth}C{0.09\linewidth} @{\hspace{10\tabcolsep}} C{0.09\linewidth}C{0.09\linewidth}C{0.09\linewidth}C{0.09\linewidth}p{0pt}}
\cline{1-9}
Method & \multicolumn{4}{c}{\quad Region $\rightarrow$ Sentence} & \multicolumn{4}{c}{\quad Sentence $\rightarrow$ Region} &  \\
& \scriptsize R@10$\uparrow$& \scriptsize R@50$\uparrow$& \scriptsize R@100$\uparrow$& \scriptsize MedR$\downarrow$ & \scriptsize R@10$\uparrow$& \scriptsize R@50$\uparrow$& \scriptsize R@100$\uparrow$& \scriptsize MedR$\downarrow$ &  \\
\cline{1-9}
GLoRIA & 0.06 & 0.21 & 0.37 & 162 & 0.06 & 0.21 & 0.34 & 183 &  \\
BioViL & 0.07 & 0.26 & 0.40 & 151 & 0.08 & 0.26 & 0.40 & 146 &  \\
LSE$+$NL & \textBF{0.11} & \textBF{0.29} & \textBF{0.45} & \textBF{119} & \textBF{0.11} & \textBF{0.36} & \textBF{0.51} & \textBF{97} &  \\
\cline{1-9}
\end{tabular}
\Ttablebottomvspace
\end{table}

{\bfseries Comparison with State-of-the-art Methods} We compare the proposed approach LSE$+$NL with the state-of-the-art methods GLoRIA \cite{huangGLoRIAMultimodalGlobalLocal2021} and BioViL \cite{boeckingMakingMostText2022}. GLoRIA is a representation learning method that learns based on image-sentence and region-word pairs. BioViL improves upon GLoRIA by using a better text encoder, relying on a symmetric contrastive loss and masked language modeling for representation learning.  We omit reporting GLoRIA's classification and visual grounding performance for GLoRIA as \cite{boeckingMakingMostText2022} showed that BioViL is better than GLoRIA on these tasks. Our simple model provides consistently better performance than these state-of-the-art algorithms.

Table~\ref{tab:exp_cmp_with_sota_rsna} reports image classification accuracy based on the learned representations for different amounts of data used to fine-tune the representation for the downstream task (zero-shot, 1\%, and 100\%). Our method is competitive or better than the baseline, especially in the zero-shot setup, underscoring its promise for limited annotation scenarios. Table~\ref{tab:exp_cmp_with_sota_grounding} and Table~\ref{tab:exp_cmp_with_sota_retrieval} report the methods' performance on visual grounding and cross-modal retrieval respectively. Our method significantly outperforms the baseline.

\begin{figure}[ht!]
    \centering
    \includegraphics[width=\textwidth]{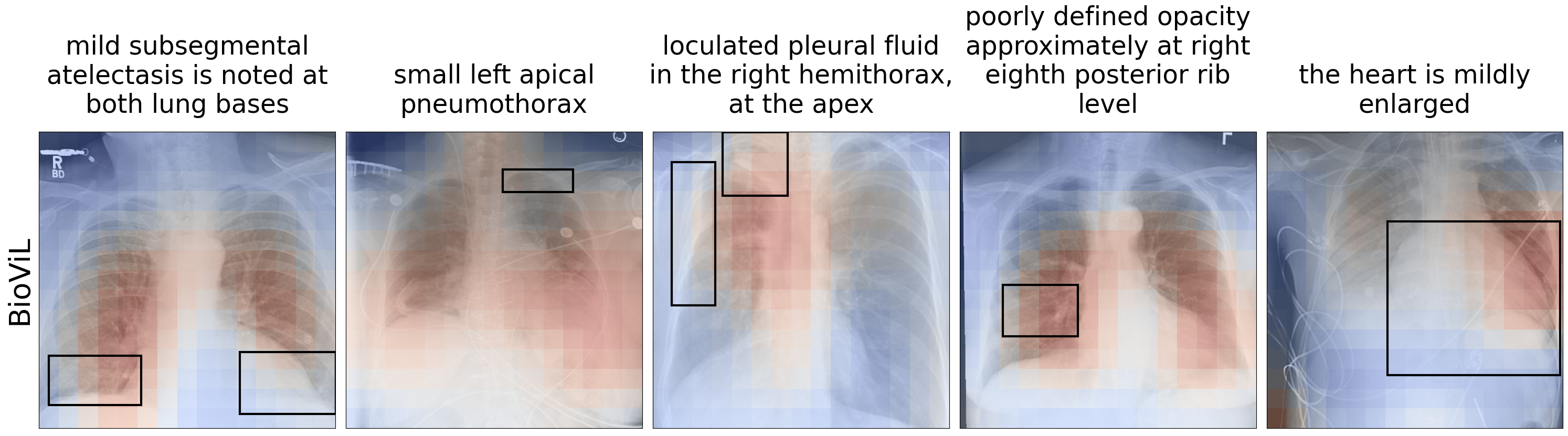}
    \includegraphics[width=\textwidth]{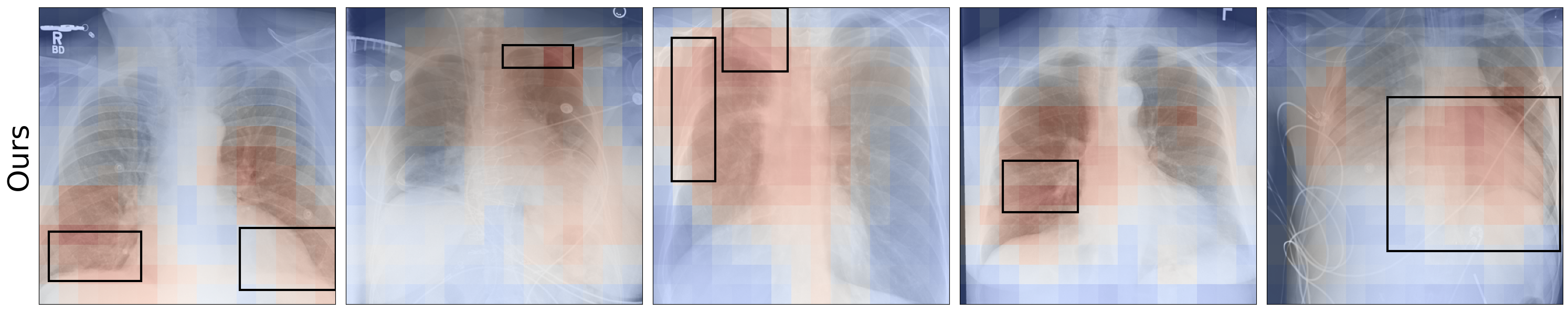}
    \caption{\footnotesize 
    Example visual grounding results for several challenging cases for BioVil \cite{boeckingMakingMostText2022} (top row) and our method (bottom row). Text queries and the corresponding ground truth bounding boxes are shown for each image. Colormap overlay visualizes region-sentence scores (blue corresponds to low scores, red highlights regions with high scores). Our method provides maps that align better with the ground truth bounding boxes. }
    \label{fig:ms_cxr_grounding_FigA2_comparisons}
    \Tfigurebottomvspace
\end{figure}

Figure~\ref{fig:ms_cxr_grounding_FigA2_comparisons} illustrates examples of visual grounding. Unlike \cite{boeckingMakingMostText2022}, we do not smooth the region-sentence scores produced by our model. Our method yield qualitatively better region-sentence scores than BioViL on a few challenging failure cases discussed in \cite{boeckingMakingMostText2022}. In particular, our pre-trained model captures location specifications more effectively, e.g., recognizing ``at both lung bases'' in the first image and ``right'' in the third image. Both our method and BioViL are prone to false positives, i.e., regions outside the ground-truth bounding box with high region-sentence scores, which highlights the need for further improvements.

\renewcommand{\arraystretch}{1}
\begin{table}[ht!]
\centering
\caption{\footnotesize Ablation study results. For each variant of the method, performance statistics are reported for each downstream task consistently with Tables \ref{tab:exp_cmp_with_sota_rsna}, \ref{tab:exp_cmp_with_sota_grounding}, and \ref{tab:exp_cmp_with_sota_retrieval}. RSNA is RSNA Pneumonia. CheX is MIMIC-CheXpert. FT is fine-tuned classification using 100\% of the labels. ZS is zero-shot classification. We report AUC for image classification. Local representations perform well for image classification, while visual grounding and cross-modal retrieval benefit from integration of local and global representations. }
\label{tab:ablation}
\begin{tabular}{L{0.14\linewidth}C{0.12\linewidth}C{0.12\linewidth}C{0.12\linewidth}|C{0.13\linewidth}|C{0.15\linewidth}C{0.15\linewidth}p{0pt}}
\cline{1-7}
Method & \multicolumn{3}{c|}{Classification} & \multicolumn{1}{c|}{Grounding} & \multicolumn{2}{c}{Cross-Modal Retrieval} & \\
& \scriptsize RSNA-ZS$\uparrow$ & \scriptsize RSNA-FT$\uparrow$ & \scriptsize CheX-FT$\uparrow$ & \scriptsize CNR$\uparrow$ & \scriptsize MedR(I \ArbitraryLengthArrow{0.2cm}T)$\downarrow$ & \scriptsize MedR(T \ArbitraryLengthArrow{0.2cm}I)$\downarrow$ &  \\
\cline{1-7}
LSE & \textBF{0.856} & \textBF{0.892} & \textBF{0.874} & 1.308 & 146 & 137 &  \\
NL & 0.636 & 0.871 & 0.854 & 0.836 & 264 & 272 &  \\
LSE+Average & 0.851 & 0.889 & 0.868 & 0.915 & 191 & 161 &  \\
LSE+NL & 0.846 & 0.891 & 0.870 & 1.403 & \textBF{110} & 102 &  \\
\scriptsize\;w. ResNet-50 & 0.844 & 0.890 & 0.870 & \textBF{1.438} & 119 & \textBF{97} &  \\
\cline{1-7}
\end{tabular}
\Ttablebottomvspace
\end{table}

\begin{figure}[ht!]
    \centering
    \includegraphics[width=\textwidth]{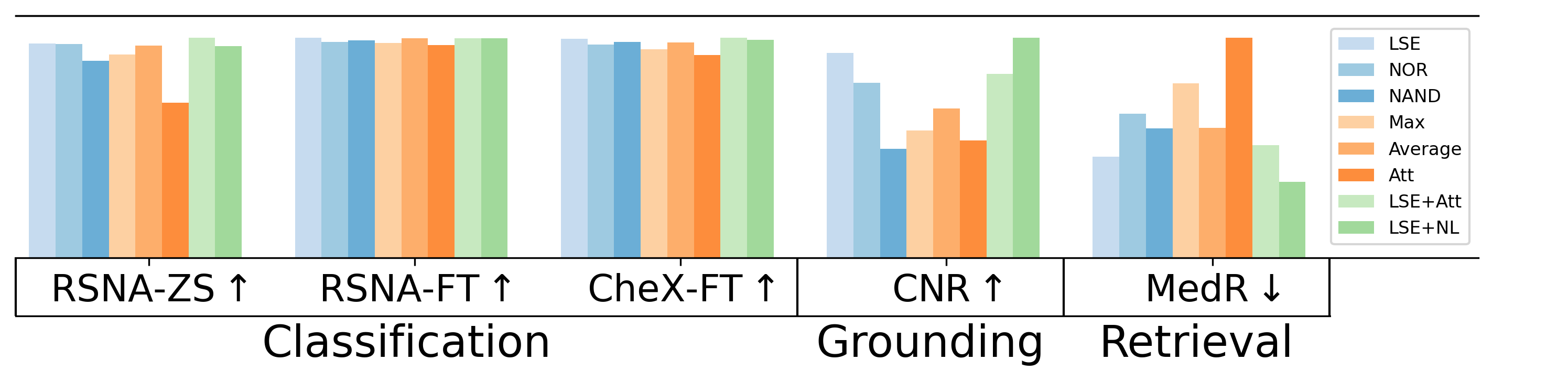}
    \caption{\footnotesize
        Effects of aggregator choice on the performance. Performance of models trained with local aggregators (shades of blue), global aggregators (shades of orange) and combinations of local and global aggregators (shades of green) is shown for image classification (AUC), visual grounding (CNR) and cross-modality retrieval (MedR averaged for both directions). The metrics are normalized to unit interval for easier comparisons across tasks. The choice of aggregators effects image classification performance much less than that of visual grounding and cross-modality retrieval. There is high performance variations within each group. Combination approaches do well on all tasks.
}
    \label{fig:cmp_instance_aggregators}
    \Tfigurebottomvspace
\end{figure}

\noindent{\bfseries Ablation} In the ablation study (Table~\ref{tab:ablation}), we compare our method LSE$+$NL with using either the local LSE or the global NL approach only, as well as replacing the NL with average as the region aggregator, i.e., LSE$+$Average. To enable extensive experimentation, we use ResNet-18 as the image encoder. LSE$+$NL provides good trade-off between region-sentence and image-sentence alignment. LSE$+$NL has comparable performance to LSE for image classification tasks while significantly outperforming all alternatives in visual grounding and cross-modal retrieval. Using a larger image encoder model ResNet-50 provides only a modest improvement in visual grounding.

\noindent{\bfseries Aggregator Choices} Figure~\ref{fig:cmp_instance_aggregators} compares a few instance aggregators' performance on downstream tasks. We compare the local approach (e.g., LSE, NOR \cite{maronFrameworkMultipleInstanceLearning1998}, NAND \cite{krausClassifyingSegmentingMicroscopy2016a}) the global approach (e.g., Max, Average, Att \cite{ilseAttentionbasedDeepMultiple2018}) and a combination of local and global approaches (e.g., LSE$+$Att, LSE$+$NL).  Aggregators within each approach exhibits high performance variations. The best local aggregator is superior to the best global aggregators we explored on all downstream tasks. Combining local and global approaches yields the best performing method.

\subsection{Limitations}

Though empirically useful, our framework doesn't provide theoretical guarantees on downstream task performance. We did not investigate what properties of an aggregator determine its transfer behaviors. In addition, our proposed method LSE$+$NL is sensitive to the value of scaling parameters; Finding the optimal hyperparameters automatically is crucial for model scaling.

\section{Conclusions}

In this paper, we propose a framework to construct permutation-invariant image-document score functions for multimodal contrastive learning. Taking inspiration from multiple instance learning, we introduce LSE$+$NL for learning multimodal representations that rely on both local and global score functions and exploit correlation between image regions. Our method outperforms the state-of-the-art approaches on image classification, visual grounding, and cross-modal retrieval. In addition, we show that contrastive representation learning is a form of multiple instance learning, providing us with valuable insights from a related field for solving shared challenges to learn representations that generalized well.

\subsubsection{Acknowledgements} Work supported by MIT JClinic, Philips, and Wistron.

\newpage
\bibliographystyle{splncs04}
\bibliography{all}

\end{document}